\documentclass[final]{cvpr}
\usepackage{times}
\usepackage{epsfig}
\usepackage{graphicx}
\usepackage{amsmath}
\usepackage{amssymb}
\usepackage{dsfont}
\usepackage{booktabs}
\usepackage{microtype}
\usepackage{tabularx}

\usepackage[pagebackref=true,breaklinks=true,colorlinks,bookmarks=false]{hyperref}

\def\cvprPaperID{4337} 
\def\confYear{CVPR 2021}

\begin{document}

\title{Multi-Objective Interpolation Training for Robustness to Label Noise}

\author{Diego Ortego, Eric Arazo, Paul Albert, Noel E. O'Connor and Kevin McGuinness \\
 Insight Centre for Data Analytics, Dublin City University (DCU)\\
 \texttt{\small{}{}\{diego.ortego, eric.arazo\}@insight-centre.org} 
}

\maketitle

\begin{abstract}

Deep neural networks trained with standard cross-entropy loss memorize noisy labels, which degrades their performance. Most research to mitigate this memorization proposes new robust classification loss functions. Conversely, we propose a Multi-Objective Interpolation Training (MOIT) approach that jointly exploits contrastive learning and classification to mutually help each other and boost performance against label noise. We show that standard supervised contrastive learning degrades in the presence of label noise and propose an interpolation training strategy to mitigate this behavior. We further propose a novel label noise detection method that exploits the robust feature representations learned via contrastive learning to estimate per-sample soft-labels whose disagreements with the original labels accurately identify noisy samples. This detection allows treating noisy samples as unlabeled and training a classifier in a semi-supervised manner to prevent noise memorization and improve representation learning. We further propose MOIT+, a refinement of MOIT by fine-tuning on detected clean samples. Hyperparameter and ablation studies verify the key components of our method. Experiments on synthetic and real-world noise benchmarks demonstrate that MOIT/MOIT+ achieves state-of-the-art results. Code is available at \url{https://git.io/JI40X}.

\end{abstract}

\section{Introduction\label{sec:Introduction}}

Building a new dataset usually involves manually labeling every sample for the particular task at hand. This process is cumbersome and limits the creation of large datasets, which are usually necessary for training deep neural networks (DNNs) in order to achieve the required performance. Conversely, automatic data annotation based on web search and user tags \cite{2017_arXiv_WebVision,2020_ICML_DatasetOOD} leverages the use of larger data collections at the expense of introducing some incorrect labels. This label noise degrades DNN performance \cite{2017_ICML_Memorization,2017_ICLR_Rethinking} and this poses an interesting challenge that has recently gained a lot of interest in the research community~\cite{2018_CVPR_IterativeNoise,2018_CVPR_JointOpt,2018_ICML_MentorNet,2019_CVPR_JointOptimizImproved,2019_ICCV_PrototypesBootstrapping,2019_ICML_DynamicBootstrapping,2020_ICLR_DivideMix,2020_CVPR_EffectiveSupervision,2020_ICML_NNweightsNoise,2020_ICML_NormalizedLossFunc}.

In image classification problems, label noise usually involves different noise distributions \cite{2020_ICML_DatasetOOD,2020_CVPR_EffectiveSupervision}. In-distribution noise types consist of samples with incorrect labels, but whose image content belongs to the dataset classes. When in-distribution noise is synthetically introduced, it usually follows either an asymmetric or symmetric random distribution. The former involves label flips to classes with some semantic meaning, e.g., a cat is flipped to a tiger, while the latter does not. Furthermore, web label noise types are usually dominated by out-of-distribution samples where the image content does not belong to the dataset classes. Recent studies show that all label noise types impact DNN performance, although performance degrades less with web noise \cite{2020_ICML_DatasetOOD,2020_ICPR_SSLnoise}.

Robustness to label noise is usually pursued by identifying noisy samples to: reduce their contribution in the loss \cite{2018_ICML_MentorNet,2018_NeurIPS_CoTeaching}, correct their label \cite{2019_ICML_DynamicBootstrapping,2020_ICLR_DivideMix}, or abstain their classification \cite{2019_ICML_AbstentionLabelNoise}. Other methods exploit interpolation training \cite{2018_ICLR_mixup}, regularizing label noise information in DNN weights \cite{2020_ICML_NNweightsNoise},
or small sets of correctly labeled data \cite{2018_NIPS_GoldLoss,2020_CVPR_EffectiveSupervision}. However, most previous methods rely exclusively on classification losses and little effort has being directed towards incorporating similarity learning frameworks \cite{2020_ECCV_MetricLearningReality}, i.e.~directly learning image representations rather than a class mapping \cite{2018_CVPR_IterativeNoise}.

Similarity learning frameworks are very popular in computer vision for a variety of applications including face recognition \cite{2018_CVPR_Cosface}, fine-grained retrieval \cite{2019_ICCV_SoftTripleLoss}, or visual search \cite{2019_ICCV_MetricLearningExample}.
These methods learn representations for samples of the same class (positive samples) that lie closer in the feature space than those of samples from different classes (negative samples). Many traditional methods are based on sampling pairs or triplets to measure similarities \cite{2005_CVPR_SiamesePairs,2018_arXiv_DeepTriplets}. However, supervised and unsupervised contrastive learning approaches that consider a high number of negatives have recently received significant attention due to their success in unsupervised learning \cite{2020_ICML_SimCLR,2020_CVPR_MoCo,2020_IEEE_COntrastiveSurvey}. In the context of label noise, there are some attempts at training with simple similarity learning losses \cite{2018_CVPR_IterativeNoise}, but there are, to the best of our knowledge, no works exploring more recent contrastive learning losses \cite{2020_arXiv_SupContLearn}.

This paper proposes Multi-Objective Interpolation Training (MOIT), a framework to robustly learn in the presence of label noise by jointly exploiting synergies between contrastive and semi-supervised learning. The former introduces a regularization of the contrastive loss in \cite{2020_arXiv_SupContLearn} to learn noise-robust representations that are key for accurately detecting noisy samples and, ultimately, for semi-supervised learning. The latter performs robust image classification and boosts performance.
Our MOIT+ refinement further demonstrates that fine-tuning on the detected clean data can boost performance. MOIT/MOIT+ achieves state-of-the-art results across a variety of datasets (CIFAR-10/100 \cite{2009_CIFAR}, mini-ImageNet \cite{2020_ICML_DatasetOOD}, and mini-WebVision \cite{2017_arXiv_WebVision}) with both synthetic and real-world web label noise. Our main contributions are as follows:
\begin{enumerate}
\item A multi-objective interpolation training (MOIT) framework where supervised contrastive learning and semi-supervised learning help each other to robustly learn in the presence of both synthetic and web label noise under a single hyperparameter configuration.
\item An interpolated contrastive learning (ICL) loss that imposes linear relations both on the input and the contrastive loss to mitigate the performance degradation observed for the supervised contrastive learning loss in \cite{2020_arXiv_SupContLearn} when training with label noise.

\item A novel label noise detection strategy that exploits the noise-robust feature representations provided by ICL to enable semi-supervised learning. This detection strategy performs a $k$-nearest neighbor search to infer per-sample label distributions whose agreements with the original labels identify correctly labeled samples.

\item A fine-tuning strategy over detected clean data (MOIT+) that further boosts performance based on simple noise robust losses from the literature.

\end{enumerate}

\section{Related work}

We briefly review recent image classification methods aiming at mitigating the effect of label noise on DNNs and recent contrastive learning methods.

\paragraph*{Noise rate estimation}

Using a label noise transition matrix can mitigate label noise \cite{2017_CVPR_ForwardLoss,2018_NIPS_GoldLoss,2019_NeurIPS_AnchorPointsMatrix}. Patrini et al.~\cite{2017_CVPR_ForwardLoss} proposed to correct the softmax classification using a transition matrix. The estimation of this matrix is, however, challenging. The authors in \cite{2019_NeurIPS_AnchorPointsMatrix} estimate the matrix by exploiting detected noisy samples that are similar to anchor points (i.e.~highly reliable detected clean samples), while Hendrycks et al.~\cite{2018_NIPS_GoldLoss} directly use a set of clean samples.

\paragraph*{Noisy sample rejection}

Rejecting or reducing the contribution to the optimization objective of noisy samples can increase model robustness~\cite{2018_ICML_MentorNet,2018_ECCV_CurrNet,2018_CVPR_IterativeNoise,2020_ICML_DatasetOOD}. Jiang et al.~\cite{2018_ICML_MentorNet} propose a teacher-student framework where the teacher estimates per-sample weights to guide the student training. Defining per-sample weights is also exploited in \cite{2018_ECCV_CurrNet}
via an unsupervised estimation of data complexity. Nguyen et al.~\cite{2020_ICLR_SELF} iteratively refine a clean set to train on by measuring label agreements with ensembled network predictions. Cross-network disagreements and updates \cite{2018_NeurIPS_CoTeaching} lead to robust learning by training on selected clean data \cite{2019_ICML_CoTeachPlus}. Also, \cite{2020_CVPR_JoCoR} propose a loss for standard training together with cross-network consistency to select the clean samples to train on. 

\paragraph*{Noisy label correction}

Correcting noisy labels to replace or balance their influence is widely used in previous works \cite{2015_ICLR_Bootstrapping,2019_ICCV_PrototypesBootstrapping,2019_ICML_DynamicBootstrapping,2020_NeurIPS_EarlyReg}.
Bootstrapping loss \cite{2015_ICLR_Bootstrapping} correction approaches exploit a perceptual term that introduces reliance on a new label given by either the model prediction with fixed \cite{2015_ICLR_Bootstrapping} or dynamic \cite{2019_ICML_DynamicBootstrapping} importance, or class
prototypes \cite{2019_ICCV_PrototypesBootstrapping}. More recently, Liu et al.~\cite{2020_NeurIPS_EarlyReg} introduced a perceptual term that maximizes the inner product between the model output and the targets without need for per-sample weights.

\paragraph{Noisy label rejection}

Rejecting the original labels by relabeling all samples with the network predictions~\cite{2018_CVPR_JointOpt} or learned label distributions~\cite{2019_CVPR_JointOptimizImproved} mitigates the effect of label noise. Recently, several approaches perform semi-supervised learning \cite{2018_WACV_SemiSupNoise,2019_ICCV_NegativeLearning,2020_ICPR_SSLnoise} by treating detected noisy samples as unlabeled, thus rejecting their labels while exploiting the image content. Their main differences are in the noise detection mechanism: Ding et al.~\cite{2018_WACV_SemiSupNoise} exploit high certainty agreements between the network predictions and labels, Kim et al.~\cite{2019_ICCV_NegativeLearning} use high softmax probabilities after performing negative learning, and Ortego et al.~\cite{2020_ICPR_SSLnoise} look at the agreements between the original and relabeled labels using \cite{2018_CVPR_JointOpt}.

\paragraph*{Other label noise methods}

Zhang et al.~\cite{2018_ICLR_mixup} proposed an interpolation training strategy, \emph{mixup}, that greatly prevents label noise memorization and has been adopted by many other methods \cite{2019_ICML_DynamicBootstrapping,2020_ICLR_DivideMix,2020_ICML_DatasetOOD,2020_ICPR_SSLnoise,2020_NeurIPS_EarlyReg}.
Harutyunyan et al.~\cite{2020_ICML_NNweightsNoise} quantify the amount of memorized information via the Shannon mutual information between neural network weights and the vector of all training labels, and encourage this to be small. Thulasidasan
et al.~\cite{2019_ICML_AbstentionLabelNoise} add an abstention class to be predicted by noisy samples due to an abstention penalty introduced in the loss. Robust loss functions are studied in several works by jointly exploiting the benefits of mean absolute error and cross-entropy losses \cite{2018_NeurIPS_GCE}, a generalized version of mutual information insensitive to noise \cite{2019_NeurIPS_LDMI}, or \cite{2020_ICML_NormalizedLossFunc} combinations of robust loss functions that mutually boost each other. Furthermore, several strategies to prevent memorization can be exploited together and DivideMix \cite{2020_ICLR_DivideMix} is a good example as it uses interpolation training, cross-network agreements, semi-supervised learning, and label correction. 

\begin{figure*}[t]
\centering{}\includegraphics[width=0.85\textwidth]{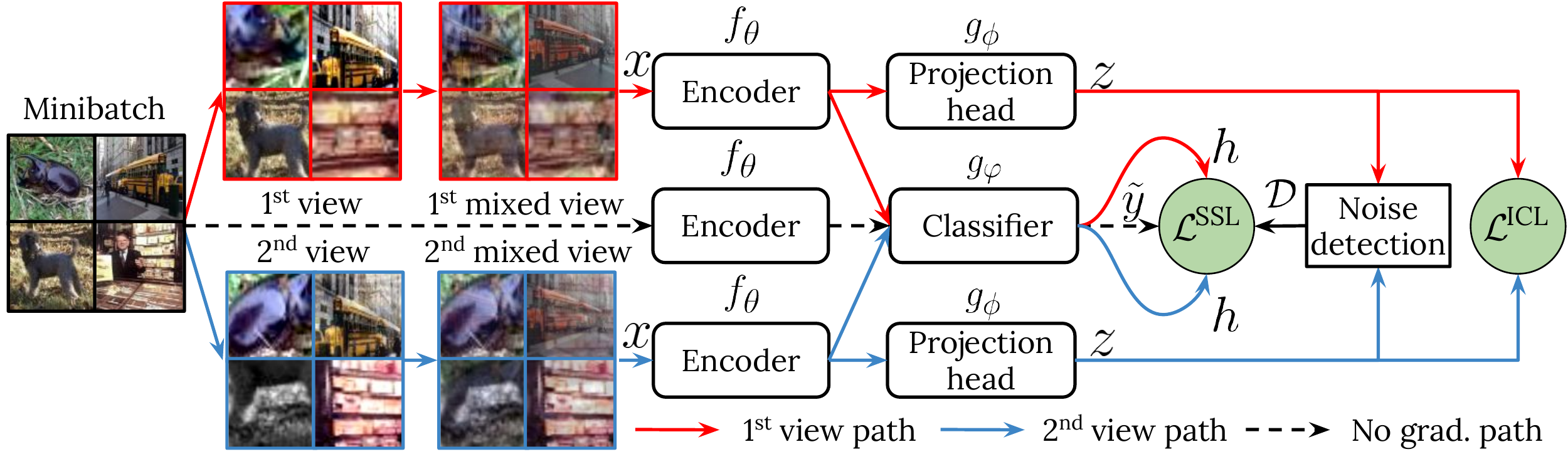}\caption{\label{fig:Overview}Multi-Objective Interpolation Training (MOIT) for improved robustness to label noise. We interpolate samples and impose the same interpolation in the supervised contrastive learning loss $\mathcal{L}^{\mathit{ICL}}$ and the semi-supervised classification loss $\mathcal{L}^{\mathit{SSL}}$ that we jointly use during training. Label noise detection is performed at every epoch to enable semi-supervised learning and its result is used after training to fine-tune the encoder and classifier to further boost performance.}
\end{figure*}

\paragraph{Contrastive representation learning}
Recent works in self-supervised learning \cite{2020_IEEE_COntrastiveSurvey, 2020_ICML_SimCLR,2020_arXiv_SupContLearn} have demonstrated the potential of contrastive based similarity learning frameworks for representation learning. These methods maximize (minimize) similarities of positive (negative) pairs. Adequate data augmentation \cite{2020_NeurIPS_GoodViews}, large amounts of negative samples via large batch size \cite{2020_ICML_SimCLR} or memory banks \cite{2020_CVPR_MoCo,2020_CVPR_XBM}, and careful network architecture designs \cite{2020_arXiv_MoCoV2} are usually important for better performance. Regarding the label noise scenario for image classification, no works explore the impact of incorrect labels on contrastive learning and only Wang et al.~\cite{2018_CVPR_IterativeNoise}  incorporate a simple similarity learning objective.


\section{Method}

We target learning robust feature representations in the presence of label noise. In particular, we adopt the contrastive learning approach from \cite{2020_arXiv_SupContLearn} and randomly sample $N$ images to apply two random data augmentation operations to each, thus generating two data views. The resulting training minibatch $\left\{ \left(x_{i},y_{i}\right)\right\} _{i=1}^{2N}$ of image-label pairs $x_{i}$ and $y_{i}$ consists of $2N$ images. Every image is mapped to a low-dimensional representation $z_{i}$ by learning an encoder network $f_{\theta}$ and a projection network $g_{\phi}$ with parameters $\theta$ and $\phi$. In particular, an intermediate embedding $v_{i}=f_{\theta}\left(x_{i}\right)$ is
generated and subsequently transformed into the representation $w_{i}=g_{\phi}\left(v_{i}\right)$. Finally, $z_{i}=w_{i}/\left\Vert w_{i}\right\Vert _{2}$ is the $L_2$-normalized low-dimensional representation used to learn based on the per-sample loss:
\begin{equation}
\mathcal{L}_{i}\left(z_{i},y_{i}\right)=\frac{1}{2N_{y_{i}}-1}\sum_{j=1}^{2N}\mathds{1}_{i\neq j}\,\mathds{1}_{y_{i}=y_{j}}P_{i,j},\label{eq:SupervisedContrastiveLoss}
\end{equation}
\begin{equation}
P_{i,j}=-\log\frac{\exp\left(z_{i}\cdot z_{j}/\tau\right)}{\sum_{r=1}^{2N}\mathds{1}_{r\neq i}\,\exp\left(z_{i}\cdot z_{r}/\tau\right)},
\end{equation}
where $P_{i,j}$ denotes the $j$-th component of the temperature $\tau$ scaled softmax distribution of inner products $z_{i}\cdot z_{j}$ of representations from the positive pair of samples $x_{i}$ and $x_{j}$, which can be interpreted as a probability. $P_{i,j}$ is aggregated in Eq. \ref{eq:SupervisedContrastiveLoss} across all $N_{y_{i}}$ samples $x_{j}$ in the minibatch sharing label with $x_{i}$ ($y_{i}=y_{j}$) except for the self-contrast case ($i=j$), as defined by the indicator function $\mathds{1}_{B}\in\left\{ 0,1\right\} $ that returns 1 when condition $B$ is fulfilled and 0 otherwise. Minimizing $\mathcal{L}_{i}$ implies adjusting $f_{\theta}$ and $g_{\phi}$ to pull together the feature representations $z_{i}$ and $z_{j}$ when they share the same label ($y_{i}=y_{j}$), while pushing them apart when they do not. Also, the gradient analysis in \cite{2020_arXiv_SupContLearn} reveals that Eq. \ref{eq:SupervisedContrastiveLoss} focuses on hard positives/negatives rather than easy ones. Note that having two data views implies that $\mathcal{L}_{i}$ contains an unsupervised contribution equivalent to the \emph{NT-Xent} loss \cite{2020_ICML_SimCLR}. 

In the presence of label noise, Eq. \ref{eq:SupervisedContrastiveLoss} incorrectly selects positive/negative samples, which degrades the feature representation\textbf{ $z$} (see Tab. \ref{tab:Weighted-k-NN-evaluation}).
To overcome this limitation and perform robust image classification under label noise conditions, we propose a Multi-Objective Interpolation Training (MOIT) framework that consists of: \emph{i)} a regularization technique to prevent memorization when training with the supervised contrastive learning loss (Sec. \ref{subsec:ICL}), \emph{ii)} a semi-supervised classification strategy based on a novel label noise detection strategy that exploits the noise-robust representation $z$ to measure agreements with the original labels $y$ and tag noisy samples as unlabeled (Sec. \ref{subsec:Semi-Supervised-Classification}), and \emph{iii)} a classifier refinement on clean data to boost classification performance (Sec. \ref{subsec:Classification-refinement}). Fig. \ref{fig:Overview} shows an overview of MOIT.

\subsection{Interpolated Contrastive Learning\label{subsec:ICL}}

Interpolation training strategies have demonstrated excellent performance
in classification frameworks \cite{2018_ICLR_mixup,2018_ACMl_RICAP,2019_ICCV_CutMix}, and have further shown promising results to prevent label noise memorization \cite{2018_ICLR_mixup,2019_ICML_DynamicBootstrapping,2020_ICLR_DivideMix,2020_ICML_DatasetOOD}. Inspired by this success, we propose Interpolated Contrastive Learning (ICL), a novel adaptation of mixup data augmentation \cite{2018_ICLR_mixup} for supervised contrastive learning. ICL performs convex combinations of pairs of samples as
\begin{equation} x_{i}=\lambda x_{a}+\left(1-\lambda\right)x_{b},\label{eq:MixupInput} \end{equation}
where $\lambda\in\left[0,1\right]\sim \mathrm{Beta}\left(\alpha,\alpha\right)$ and $x_{i}$ denotes the training sample that combines two minibatch samples $x_{a}$ and $x_{b}$, and imposes a linear relation in the contrastive loss:  
\begin{equation}
\mathcal{L}_{i}^{\mathit{MIX}}=\lambda\mathcal{L}_{i}\left(z_{i},y_{a}\right)+\left(1-\lambda\right)\mathcal{L}_{i}\left(z_{i},y_{b}\right).\label{eq:ContrastiveMix}
\end{equation}
The first and second terms in Eq. \ref{eq:ContrastiveMix} consider, respectively, positive samples from the class $y_{a}$ ($y_{b}$) given by the first (second) sample $x_{a}$ ($x_{b}$). The selection of positive/negative samples involves considering a unique class for every mixed example. However, in most cases the input samples contain two classes as a result of the interpolation, where $\lambda$ determines the dominant one. We assign this dominant class to every sample for positive/negative sampling. Intuitively, ICL makes it harder to pull together clean and noisy samples with the same label, as noisy samples are interpolated with either another clean sample that provides a
clean pattern beneficial for training or another noisy sample that makes it harder to memorize the noisy pattern. 

\paragraph{Memory bank}

The number of positives and negatives selected for contrastive learning depends on the minibatch size and the number of dataset classes. Therefore, unless a large minibatch is used during training, few positive and negative samples are selected, which negatively affects the training process \cite{2020_arXiv_SupContLearn}. To address limitations in computing resources, we introduce the memory bank proposed in \cite{2020_CVPR_XBM} to perform robust similarity learning despite using relatively small minibatches compared to those in \cite{2020_arXiv_SupContLearn}. In particular, we define a memory to store the last $M$ feature representations from previous minibatches and define a loss term $\mathcal{L}_{i}^{\mathit{MEM}}$ similar to $\mathcal{L}_{i}^{\mathit{MIX}}$ in Eq. \ref{eq:ContrastiveMix}. While $\mathcal{L}_{i}^{\mathit{MIX}}$ is estimated contrasting the $2N$ minibatch samples across them, $\mathcal{L}_{i}^{\mathit{MEM}}$ contrasts the $2N$ samples with the $M$ memory samples, thus extending the number of positive and negative samples. The final ICL loss then aggregates the average batch and memory losses:
\begin{equation}
\mathcal{L}^{\mathit{ICL}}=\mathcal{L}^{\mathit{MIX}}+\mathcal{L}^{\mathit{MEM}}.
\end{equation}
Sec. \ref{subsec:ICLexperiments} shows the benefits of using ICL loss instead of the original loss in \cite{2020_arXiv_SupContLearn} and Sec. \ref{subsec:Joint-training-hyperparameters} demonstrates the effect of the memory bank on the overall method.

\subsection{Semi-Supervised Classification\label{subsec:Semi-Supervised-Classification}}

The goal is to predict a class $c\in\left\{ 1,\ldots,C\right\} $ by learning a second mapping $h(v)=g_{\varphi}\left(v\right)$ to the class space, where $C$ is the number of classes. Na\"ively training a classifier in the presence of label noise leads to noise memorization \cite{2017_ICML_Memorization,2017_ICLR_Rethinking}, which degrades the performance.
Semi-supervised learning, where noisy labels are discarded, can mitigate this memorization \cite{2018_WACV_SemiSupNoise,2020_ICLR_DivideMix,2020_ICPR_SSLnoise}. We, therefore, propose to jointly adopt semi-supervised learning with ICL. The former boosts the performance achievable by the latter, while the latter enables accurate label noise detection necessary for good performance in the former.

\paragraph*{Label noise detection }

We propose to measure agreements between the feature representation $z_{i}$ (robust to label noise) and the original label $y_{i}$ to identify mislabeled samples. To quantify this agreement, we start by estimating a class probability distribution from the representation $z_{i}$ by doing a $k$-nearest neighbor (k-NN) search:
\begin{equation}
p\left(c\mid x_{i}\right)=\frac{1}{K}\sum_{\underset{x_{k}\in\mathcal{N}_{i}}{k=1}}^{K}\mathds{1}_{y_{k}\neq c},\label{eq:kNN_Initial}
\end{equation}
where $\mathcal{N}_{i}$ denotes the neighbourhood of $K$ closest images to $x_{i}$ according to the feature representation $z$. Eq. \ref{eq:kNN_Initial} then counts the number of samples per class in the local neighborhood $\mathcal{N}_{i}$ and normalizes the counts to estimate a probability distribution. This distribution can be interpreted as a soft-label that can be compared with the original label to identify potential disagreements, i.e.~noisy samples. However, the labels $y$ might be noisy, thus biasing the estimation of $p$. We, therefore, estimate a corrected distribution $\hat{p}$ using:
\begin{equation} 
\hat{p}\left(c\mid x_{i}\right)=\frac{1}{K}\sum_{\underset{x_{k}\in\mathcal{N}_{i}}{k=1}}^{K}\mathds{1}_{\hat{y}_{k}\neq c},\label{eq:kNN_final}
\end{equation}
where we introduce corrected labels $\hat{y}$ that are estimated taking the dominant label in $\mathcal{N}_{i}$, i.e.~$\hat{y}=\arg\max_{c}\:p\left(c\mid x\right)$. Finally, the disagreement between the corrected distribution $\hat{p}\left(c\mid x_{i}\right)$ and the label noise distribution given by the original label $y_{i}$ is measured by the cross-entropy
\begin{equation}
d_{i}=-y_{i}^{T}\log\left(\hat{p}\right),
\end{equation}
where $T$ denotes the transpose operation. The higher $d_{i}$, the higher the disagreement between distributions and the more likely $x_{i}$ is a noisy sample. We select clean samples for each class $c$ based on $d_{i}$ using: 
\begin{equation}
\mathcal{D}_{c}=\left\{ \left(x_{i},y_{i}\right):d_{i}\leq\gamma_{c}\right\} ,
\end{equation}
where $\gamma_{c}$ is a per-class threshold on $d_{i}$, which is dynamically
defined to ensure a balanced clean set across classes. To perform this balancing, we use the median of per-class agreements between the corrected label $\hat{y}_{i}$ and the original label $y_{i}$ across all classes. Sec. \ref{subsec:Joint-training-hyperparameters} illustrates the importance of this balancing strategy as well as the corrected distribution $\hat{p}$ over $p$ for achieving better performance. Note that a k-NN noise detection that resembles Eq. \ref{eq:kNN_Initial} has been recently proposed in \cite{2020_ICML_DeepKNN}. However, we differ in that we propose a corrected version in Eq. \ref{eq:kNN_final} that surpasses the straightforward k-NN of Eq. \ref{eq:kNN_Initial} (see Tab. \ref{tab:NoiseDetStudy}), we use k-NN during training, and always avoid using a trusted clean set.

\paragraph*{Semi-supervised learning}

We learn the classifier by performing semi-supervised learning where samples in $\mathcal{D}$ are considered as labeled and the remaining samples as unlabeled. To leverage these unlabeled samples, pseudo-labeling \cite{2020_IJCNN_Pseudo} based on interpolated samples is applied by defining the objective
\begin{equation}
\mathcal{L}_{i}^{\mathit{SSL}}=-\lambda\tilde{y}_{a}^{T}\log\left(h_{i}\right)-\left(1-\lambda\right)\tilde{y}_{b}^{T}\log\left(h_{i}\right),\label{eq:SSL}
\end{equation}
where the pseudo-label $\tilde{y}_{a}$ ($\tilde{y}_{b}$) for
$x_{a}$ ($x_{b}$) is estimated as
\begin{equation}
\tilde{y}_{a}=\begin{cases}
y_{a}, & x_{a}\in\mathcal{D}_{c}\\
\bar{h}_{a}, & x_{a}\notin\mathcal{D}_{c}
\end{cases},
\end{equation}
where $\bar{h}_{a}$ is the softmax prediction for image $x_{a}$ without data augmentation. The final Multi-Objective Interpolation Training (MOIT)  optimizes the loss:
\begin{equation}
\mathcal{L}^{\mathit{MOIT}}=\mathcal{L}^{\mathit{ICL}}+\mathcal{L}^{\mathit{SSL}}.\label{eq:FinalLoss}
\end{equation}
In summary, the proposed MOIT framework enables robust training in the presence of label noise by learning robust representations via contrastive learning that help in achieving successful noise detection that discards noisy labels and enables semi-supervised learning for classification. Note that the method needs to learn useful features before performing accurate noise detection; thus we start training with $\tilde{y}=y,\forall x$ in $\mathcal{L}^{\mathit{SSL}}$, i.e.~a normal supervised training. We start doing semi-supervised learning once reasonable features to search for reliable nearest neighbors in Eq. \ref{eq:kNN_final} are learned and the clean sample detection is made reliable. We assume that good features are available soon after reducing the learning rate, given that there is little risk of overfitting noisy labels at earlier epochs when using a high learning rate, as often reported in the literature
\cite{2018_CVPR_JointOpt,2019_CVPR_JointOptimizImproved,2019_ICML_DynamicBootstrapping}.

\subsection{Classification refinement\label{subsec:Classification-refinement}}

Supervised pre-training on relatively clean datasets such as ImageNet \cite{2009_CVPR_ImageNet} has proved to mitigate label noise memorization \cite{2019_ICML_PreTraining,2020_ICML_DatasetOOD}. We, therefore, refine our MOIT predictions by fine-tuning $f_{\theta}$ and re-training $g_{\varphi}$ on our detected clean set $\mathcal{D}$ using a constant low learning rate. We name this fine-tuning stage MOIT+. We train using mixup \cite{2018_ICLR_mixup} and later introduce  hard bootstrapping loss correction \cite{2015_ICLR_Bootstrapping} to deal with possible low amounts of label noise present in $\mathcal{D}$, thus defining the following training objective: 
\begin{equation*}
\mathcal{L}_{i}^{\mathit{MOIT+}}=-\lambda\left[\left(\delta y_{a}+\left(1-\delta\right)\tilde{y}_{a}\right)^{T}\log\left(\mathit{h_{i}}\right)\right]-
\end{equation*}
\begin{equation}
\left(1-\lambda\right)\left[\left(\delta y_{b}+\left(1-\delta\right)\tilde{y}_{b}\right)^{T}\log\left(\mathit{h_{i}}\right)\right],\label{eq:LossMOIT+}
\end{equation}
where $\lambda$ is the mixing coefficient from \cite{2018_ICLR_mixup} as we are interpolating images as explained in Eq. \ref{eq:MixupInput}, and $\delta$ is a weight to balance the contribution of the original labels ($y_{a}$ and $y_{b}$) or the network predictions ($\tilde{y}_{a}$ and $\tilde{y}_{b}$). This training objective is similar to that used in \cite{2019_ICML_DynamicBootstrapping}, but different in that we do not train from scratch using all data, or need to infer per sample $\delta$ weights. Instead, we set $\delta=0.8$ as done in \cite{2015_ICLR_Bootstrapping} to give more importance to the original labels, which is reasonable given that the training uses the detected clean data $\mathcal{D}$. Note that $\tilde{y}_{a}=\arg\max_{c}\:\bar{h}_{a}$ ($\tilde{y}_{b}=\arg\max_{c}\:\bar{h}_{b}$) is the network prediction for $x_{a}$ ($x_{b}$) without data augmentation. As commented before, MOIT+ starts with a mixup training without bootstrapping (i.e. $\delta=1.0$) during the initial epochs to allow adequate re-training of $g_{\varphi}$ before trusting its predictions.

\section{Experiments}

We first run experiments on the standard benchmarks for synthetic noise in CIFAR-100 \cite{2009_CIFAR} aiming at analyzing the different components of our method. We further perform comparative evaluations against related work using synthetic label noise in CIFAR-10/100, controlled web noise in mini-ImageNet \cite{2020_ICML_DatasetOOD}, and the uncontrolled web noise from the WebVision dataset \cite{2017_arXiv_WebVision}. 

\subsection{Datasets}

The CIFAR-10/100 datasets \cite{2009_CIFAR} contain 50K (10K) small resolution images for training (test). For hyperparameter and ablation studies, we keep 5K training samples for validation using their correct labels. However, to facilitate comparison with related work, we train with the full 50K samples and use the 10K test set for evaluation (reporting accuracy in the last epoch). For noise addition, we follow the criteria in \cite{2019_CVPR_JointOptimizImproved}: symmetric noise is introduced by randomly flipping the labels of a percentage of the training set to incorrect labels; asymmetric noise uses label flips to incorrect classes ``truck $\rightarrow$ automobile, bird $\rightarrow$ airplane, deer $\rightarrow$ horse, cat $\rightarrow$ dog'' in CIFAR-10, whereas in CIFAR-100 label flips are done circularly within the super-classes.


Jiang et al.~\cite{2020_ICML_DatasetOOD} propose to use mini-ImageNet and Stanford Cars to introduce both web and symmetric in-distribution
noise in a controlled manner with different noise ratios. We adopt the mini-ImageNet web noise dataset for evaluation in a real scenario with several ratios, which consists of 100 classes with 50K (5K) samples for training (validation). For further evaluation against web noise, we adopt the mini-WebVision dataset \cite{2020_ICLR_DivideMix} that uses the top-50 classes from the Google image subset of WebVision \cite{2017_arXiv_WebVision}.

\subsection{Training details}

\begin{table}[!t]
\centering{}\caption{\label{tab:Training-details}Training details. We always train from
scratch. LR: Learning rate. B: Bootstrapping.}
\smallskip{}
\setlength{\tabcolsep}{3pt}{\small{}}%
\begin{tabular}{llll}
\toprule
 & {\small{}CIFAR} & {\small{}mini-ImageNet} & {\small{}mini-WebVision}\tabularnewline
\midrule
{\small{}Resolution} & {\small{}$32\times32$} & {\small{}$84\times84$} & {\small{}$224\times224$}\tabularnewline

{\small{}Batch size} & {\small{}128} & {\small{}64} & {\small{}64}\tabularnewline

{\small{}Mem. size} & {\small{}20K} & {\small{}100K} & {\small{}50K}\tabularnewline

{\small{}Network} & {\small{}PRN-18} & {\small{}RN-18} & {\small{}RN-18}\tabularnewline

{\small{}Epochs} & {\small{}250} & {\small{}130} & {\small{}130}\tabularnewline

{\small{}Optimizer} & \multicolumn{3}{l}{{\small{}SGD, momentum 0.9, weight decay $10^{-4}$ }}\tabularnewline

{\small{}Initial LR} & {\small{}0.1} & {\small{}0.1} & {\small{}0.1}\tabularnewline

{\small{}LR decay} & {\small{}125, 200} & {\small{}80, 105} & {\small{}80, 105}\tabularnewline

{\small{}Decay factor} & {\small{}$\times$0.1} & {\small{}$\times$0.1} & {\small{}$\times$0.1}\tabularnewline

{\small{}SSL epoch} & {\small{}130} & {\small{}85} & {\small{}85}\tabularnewline

{\small{}Decay factor} & {\small{}$\times$0.1} & {\small{}$\times$0.1} & {\small{}$\times$0.1}\tabularnewline

{\small{}Epochs (MOIT+)} & {\small{}70} & {\small{}50} & {\small{}50}\tabularnewline

{\small{}LR (MOIT+)} & \multicolumn{3}{c}{{\small{}0.001 (not reduced)}}\tabularnewline

{\small{}B epoch (MOIT+)} & {\small{}30} & {\small{}20} & {\small{}20}\tabularnewline
\bottomrule
\end{tabular}
\end{table}

We use a PreAct ResNet-18 (PRN-18) \cite{2016_ECCV_PreActResNet} as encoder network in CIFAR following \cite{2019_ICML_DynamicBootstrapping}, while for mini-ImageNet we use the ResNet-18 (RN-18) from \cite{2020_arXiv_ResNetFewShot} used in mini-ImageNet for few-shot learning. For mini-WebVision we use a standard RN-18 \cite{2016_CVPR_ResNet}. We do not evaluate using other frameworks in mini-ImageNet or WebVision \cite{2020_ICML_DatasetOOD,2020_ICLR_DivideMix} due to limitations of our computing resources. We, conversely, re-run the official implementation of top-performing and recent methods \cite{2018_ICLR_mixup,2020_ICLR_DivideMix,2020_NeurIPS_EarlyReg} in our framework. As projection head and classifier, we always use a linear layer that maps, respectively, to a feature vector $z$ of 128 dimensions and to the class space.

Table \ref{tab:Training-details} presents the training details for MOIT and MOIT+. We interpolated input samples as proposed in \cite{2018_ICLR_mixup} with $\alpha=1$ (i.e.~$\lambda$ is sampled from a uniform distribution), and apply standard strong data augmentations to achieve successful contrastive learning\footnote{\url{https://github.com/HobbitLong/SupContrast}} in MOIT: random resized crops, horizontal flips, color jitter and gray scale transformations. For MOIT+ and all other methods, mixup as well as standard augmentations are used (CIFAR: random horizontal flips and random 4 pixel translations, mini-ImageNet and mini-WebVision: random resized crops and random horizontal flips). We double the epochs in MOIT+ for 80\% noise in CIFAR-10/100 as there are few selected clean samples, which make epochs extremely short. We always use $\tau=0.1$ temperature scaling for contrastive learning and increase the memory size in mini-ImageNet and mini-WebVision to deal with reduced batch size. Note that MOIT+ finetunes the model in the last epoch when training MOIT.

In practice, the noise ratio and distribution are not usually known a-priori; we therefore use a common configuration for training our method (mixup $\alpha$, $k$-NN parameter $K$, loss function, $\mathcal{D}$ balancing criterion, $\delta$ for MOIT+), and only modify typical hyperparameters (batch size, memory and epochs). We use the official implementations of DivideMix (DMix) \cite{2020_ICLR_DivideMix} and ELR \cite{2020_NeurIPS_EarlyReg}. However, DMix adopts specific configurations for different datasets and even for different noise ratios and types in the same dataset. To perform as fair as possible a comparison without degrading DMix results, we select a single parametrization of DMix in every dataset based on the most repeated configuration in \cite{2020_ICLR_DivideMix}. This affects the CIFAR configuration (CIFAR-10: $\lambda_{u}=0$, CIFAR-100: $\lambda_{u}=150$) as mini-WebVision has a unique configuration that we also adopt for mini-ImageNet. We run DMix and ELR for the same number of epochs as our method respecting suggested learning rates and equip ELR with \textit{mixup} for a fair comparison with DMix and our method that both use interpolation training. Note that ELR+ in \cite{2020_NeurIPS_EarlyReg} uses mixup, but we do not use it for comparison as it involves using a second network and a weight averaging.

\subsection{Supervised contrastive learning and label noise\label{subsec:ICLexperiments}}

\begin{table}[!t]
\centering{}\caption{\label{tab:Weighted-k-NN-evaluation}Weighted k-NN evaluation in CIFAR-100.}
\smallskip{}
\setlength{\tabcolsep}{4pt}{\small{}}%
\begin{tabularx}{\columnwidth}{Xlllll}
\toprule
 &  & \multicolumn{2}{c}{{\small{}Symmetric}} & \multicolumn{2}{l}{{\small{}Asymmetric}}\tabularnewline
 & {\small{}0\%} & {\small{}40\%} & {\small{}80\%} & {\small{}10\%} & {\small{}40\%}\tabularnewline
\midrule
{\small{}SCL} & {\small{}72.66} & {\small{}58.32} & {\small{}41.00} & {\small{}71.11} & {\small{}68.00}\tabularnewline

{\small{}ICL} & {\small{}75.30} & {\small{}66.38} & {\small{}53.60} & {\small{}74.34} & {\small{}72.04}\tabularnewline
 
{\small{}MOIT} & \textbf{\small{}75.76} & \textbf{\small{}67.42} & \textbf{\small{}55.58} & \textbf{\small{}74.86} & \textbf{\small{}72.60}\tabularnewline
\bottomrule
\end{tabularx}
\end{table}

\begin{table*}[t]
\begin{centering}
\caption{\label{tab:NoiseDetStudy}Classification accuracy for different noise detection strategies and $K$ values for 40\% asymmetric noise in CIFAR-100.}
\medskip{}
\par\end{centering}
\centering{}\setlength{\tabcolsep}{4pt}{\small{}}%
\begin{tabular}{llllllllllll}
\toprule
 & {\small{}$K$} & {\small{}5} & {\small{}10} & {\small{}25} & {\small{}50} & {\small{}100} & {\small{}150} & {\small{}200} & {\small{}250} & {\small{}300} & {\small{}350}\tabularnewline
\midrule
{\small{}k-NN ($p$)} & {\small{}Acc.} & {\small{}59.42} & {\small{}61.74} & {\small{}64.84} & {\small{}66.10} & {\small{}67.18} & {\small{}67.42} & {\small{}67.46} & {\small{}67.68} & {\small{}67.14} & {\small{}66.94}\tabularnewline

{\small{}k-NN ($\hat{p}$)} & {\small{}Acc.} & {\small{}62.28} & {\small{}65.30} & {\small{}68.58} & {\small{}70.56} & {\small{}71.16} & {\small{}71.22} & {\small{}71.24} & \textbf{\small{}71.42} & {\small{}70.98} & {\small{}70.80}\tabularnewline
\bottomrule
\end{tabular}{\small\par}
\end{table*}
\begin{table}[t]
\centering{}\caption{\label{tab:Effect-of-balancing}Effect on classification accuracy of the balancing strategy for  the clean set  $\mathcal{D}$ in CIFAR-100.
A: Asymmetric. S: Symmetric.}
\medskip{}
{\small{}}%
\begin{tabularx}{\columnwidth}{Xllll}
\toprule
 & {\small{}Unbalanced} & {\small{}Min} & {\small{}Max} & {\small{}Median}\tabularnewline
\midrule
{\small{}A-40\%} & {\small{}69.58} & {\small{}52.88} & {\small{}62.58} & \textbf{\small{}71.42}\tabularnewline

{\small{}S-40\%} & {\small{}66.28} & {\small{}63.26} & {\small{}66.12} & \textbf{\small{}66.58}\tabularnewline
\bottomrule
\end{tabularx}
\end{table}
We start by analyzing supervised contrastive learning behavior in the presence of label noise and how introducing interpolation training impacts the learned representations. We evaluate the quality of  representations using a \emph{weighted $k$-NN} ($k=200$) evaluation typical in unsupervised learning \cite{2019_ICML_AND}.
Tab.~\ref{tab:Weighted-k-NN-evaluation} reports this evaluation using the embedding $z$ extracted after the projection head (model from the last training epoch) and the true labels in the training set. This experiments show that Supervised Contrastive Learning (SCL) \cite{2020_arXiv_SupContLearn} performance degrades when there is label noise (the noise-free accuracy of 72.66 decreases). The proposed regularization using Interpolated Contrastive Learning (ICL) mitigates label noise drops and outperforms SCL in the noise-free case, validating the utility of imposing a interpolated behavior in the contrastive loss. Note that ICL and MOIT (joint ICL and semi-supervised classification) perform worse in the asymmetric case than in the symmetric case. This occurs due to the former having label flips that keep some semantic meaning (e.g. cat$\rightarrow$dog), while the latter does not (e.g. cat$\rightarrow$truck). Semantic noise is more informative during ICL, which leads to better performance and less room for semi-supervised learning improvement in MOIT compared to ICL.
We train SCL and ICL using a memory bank for 350 epochs with initial learning rate of 0.1, divided by 10 at epochs 200 and 300. Note that contrastive
learning frameworks tend to be very sensitive to hyperparameters \cite{2020_ICML_SimCLR,2020_CVPR_MoCo,2020_arXiv_SupContLearn}
(learning rate, temperature, data augmentation, etc.), a behavior that we also observed when training them alone in the presence of label noise. We experimentally found that averaging the contrastive losses of the minibatch $\mathcal{L}^{\mathit{MIX}}$ and the memory $\mathcal{L}^{\mathit{MEM}}$ helped convergence in SCL and ICL and used it in this experiment. Adding a classification objective, as done in the proposed MOIT method, stabilizes this behavior and achieves better representations than SCL and ICL alone (see Tab.~\ref{tab:Weighted-k-NN-evaluation}).

\subsection{Label noise detection analysis\label{subsec:Joint-training-hyperparameters}}

We exploit the feature representation $z$ by searching the closest $K$ neighbors to estimate a corrected soft-label $\hat{p}$ in Eq.~\ref{eq:kNN_final} and measure agreements with the original labels $y$. Tab. \ref{tab:NoiseDetStudy} shows that using this corrected soft-label $\hat{p}$ (bottom) rather than the soft-label $p$ from Eq. \ref{eq:kNN_Initial} (top) results in better performance due to improved label noise detection: precision and  recall for $\hat{p}$ are 90.83 and 87.84 compared to 80.20 and 84.43 for ${p}$. The method is also not very sensitive to the value of $K$ once it is set to a high enough value. We adopt $K=250$ for the remaining experiments. We further study the effect of balancing the clean set $\mathcal{D}$ (see Tab. \ref{tab:Effect-of-balancing}). In particular, we experiment by balancing with the minimum (Min), maximum (Max), or median (used by our method) number of agreements between corrected $\hat{y}$ and original $y$ labels across classes. The median consistently outperforms the others as it poses a better trade-off than the \emph{Min} (\emph{Max}), which restricts (extends) the samples to select in classes with many (few) agreements. Here the unbalanced criterion considers as clean all samples that satisfy $\hat{y}=y$.

\subsection{Joint training ablation study\label{subsec:Joint-training-ablation}}

\begin{table}[t]
\begin{centering}
\caption{\label{tab:AblationJoint}Ablation study for MOIT and MOIT+ in CIFAR-100.
A: Asymmetric, S: Symmetric, SSL: semi-supervised learning, M: memory,
B: Balanced clean set, r-t C: Re-training classifier, s-DA: strong
data augmentation.}
\medskip{}
{\small{}}{\small\par}
\par\end{centering}
\centering{}{\small{}}%
\begin{tabularx}{\columnwidth}{Xll}
\toprule 
 & {\small{}S-40\%} & {\small{}A-40\%}\tabularnewline
\midrule 
{\small{}(MOIT) w/o SSL} & {\small{}62.82} & {\small{}53.73}\tabularnewline

{\small{}(MOIT) w/o M} & {\small{}66.10} & {\small{}68.88}\tabularnewline

{\small{}(MOIT) w/o B} & {\small{}66.28} & {\small{}69.58}\tabularnewline

{\small{}MOIT} & \textbf{\small{}66.58} & \textbf{\small{}71.42}\tabularnewline
\midrule
{\small{}(MOIT+) w/o r-t C} & {\small{}69.54} & {\small{}73.32}\tabularnewline

{\small{}(MOIT+) w/ s-DA} & {\small{}67.98} & {\small{}71.90}\tabularnewline

{\small{}MOIT+} & \textbf{\small{}70.68} & \textbf{\small{}73.58}\tabularnewline
\bottomrule
\end{tabularx}{\small{}}{\small\par}
\end{table}
Tab. \ref{tab:AblationJoint} illustrates the effect of removing key components of our method on classification accuracy. Removing semi-supervised learning (SSL) involves training the classifier using \textit{mixup}, which results in substantial degradation due to label noise memorization. Removing the memory (M) decreases performance due to the limited batch size used (128), which provides few positives/negatives for supervised contrastive learning with 100 classes. Not balancing (B) the clean set $\mathcal{D}$ to perform SSL also decreases performance. The criterion used to select clean samples without balancing was to select every sample $x$ satisfying the agreement $\hat{y}=y$ as studied in Sec. \ref{subsec:Joint-training-hyperparameters}. Regarding the classifier refinement done by MOIT+, re-training the classifier (r-t C) and avoiding the use of strong data augmentation impact performance. The former might prevent some slight memorization behavior in the classifier occurring during MOIT, while the latter avoids the strong data augmentation that harms classification accuracy but is required for successful contrastive learning. 

\subsection{Synthetic label noise evaluation}

Tables \ref{tab:CIFAR10eval} and \ref{tab:CIFAR100eval} evaluate the performance of MOIT and MOIT+ in, respectively, CIFAR-10 and CIFAR-100 for different levels of symmetric and asymmetric noise and report average accuracy for each dataset to ease comparison. We compare against some relevant and recent methods from the literature \cite{2018_ICLR_mixup,2019_ICML_DynamicBootstrapping,2019_CVPR_JointOptimizImproved,2019_NeurIPS_LDMI,2020_ICLR_DivideMix,2020_NeurIPS_EarlyReg} and demonstrate that MOIT and MOIT+ achieve state-of-the-art results. We achieve especially robust results for asymmetric noise, which is more realistic than symmetric as label flips are done considering semantic similarities between classes. We run DMix (evaluation done without ensembling both networks) and ELR, while using the remaining results from \cite{2020_ICPR_SSLnoise}, which used the same network architecture and label noise criterion. DMix \cite{2020_ICLR_DivideMix} and, especially, ELR outperform our method for some noise levels, but experience important drops at high noise levels, which penalize the average performance. We stress that our label noise criterion (also adopted in \cite{2018_ICML_MentorNet, 2018_CVPR_IterativeNoise, 2020_ICPR_SSLnoise}) considers 40\% noise as 0.4 probability of flipping the label to an \textit{incorrect} class, and not to any class as reported in the DMix and ELR papers \cite{2020_ICLR_DivideMix,2020_NeurIPS_EarlyReg}, which results in 40\% being more challenging in our setup. 
\begin{table}[t]
\centering{}\caption{\label{tab:CIFAR10eval}Performance in CIFAR-10 with symmetric
and asymmetric noise. ({*}) Denotes that we have run the algorithm.}
\medskip{}
\setlength{\tabcolsep}{3pt}\resizebox{1\columnwidth}{!}{{\small{}}%
\begin{tabular}{llllllll|l}
\toprule 
 &  & \multicolumn{3}{l}{{\small{}Symmetric}} & \multicolumn{3}{l|}{{\small{}Asymmetric}} & {\small{}Avg.}\tabularnewline
 & {\small{}0\%} & {\small{}20\%} & {\small{}40\%} & {\small{}80\%} & {\small{}10\%} & {\small{}30\%} & {\small{}40\%} & \tabularnewline
\midrule 
{\small{}CE} & {\small{}93.85} & {\small{}78.93} & {\small{}55.06} & {\small{}33.09} & {\small{}88.81} & {\small{}81.69} & {\small{}76.04} & {\small{}72.50}\tabularnewline
 
{\small{}Mix \cite{2018_ICLR_mixup}} & \textbf{\small{}95.96} & {\small{}84.76} & {\small{}66.07} & {\small{}20.38} & {\small{}93.30} & {\small{}83.26} & {\small{}77.74} & {\small{}74.50}\tabularnewline

{\small{}DB \cite{2019_ICML_DynamicBootstrapping}} & {\small{}79.18} & {\small{}93.82} & {\small{}92.26} & {\small{}15.53} & {\small{}89.58} & {\small{}92.20} & {\small{}91.20} & {\small{}79.11}\tabularnewline

{\small{}DMI \cite{2019_NeurIPS_LDMI}} & {\small{}93.88} & {\small{}88.33} & {\small{}83.24} & {\small{}43.67} & {\small{}91.11} & {\small{}91.16} & {\small{}83.99} & {\small{}82.20}\tabularnewline

{\small{}PCIL \cite{2019_CVPR_JointOptimizImproved}} & {\small{}93.89} & {\small{}92.72} & {\small{}91.32} & {\small{}55.99} & {\small{}93.14} & {\small{}92.85} & {\small{}91.57} & {\small{}87.35}\tabularnewline

{\small{}DRPL \cite{2020_ICPR_SSLnoise}} & {\small{}94.08} & {\small{}94.00} & {\small{}92.27} & {\small{}61.07} & \textbf{\small{}95.50} & {\small{}92.98} & {\small{}92.84} & {\small{}88.96}\tabularnewline

{\small{}DMix{*} \cite{2020_ICLR_DivideMix}} & {\small{}94.27}  & \textbf{\small{}95.12}  & \textbf{\small{}94.11} & {\small{}35.36}  & {\small{}93.77}  & {\small{}92.47} & {\small{}90.04} & {\small{}85.02} \tabularnewline

{\small{}ELR{*} \cite{2020_NeurIPS_EarlyReg}} & {\small{}95.49}  & {\small{}94.49}  & {\small{}92.56}  & {\small{}38.23}  & {\small{}95.25}  & \textbf{\small{}94.66}  & {\small{}92.88}  & {\small{}86.22} \tabularnewline
\midrule
{\small{}MOIT} & {\small{}95.17} & {\small{}92.88} & {\small{}90.55} & {\small{}70.53} & {\small{}93.50} & {\small{}93.19} & {\small{}92.27} & {\small{}89.73}\tabularnewline

{\small{}MOIT+} & {\small{}95.65} & {\small{}94.08} & {\small{}91.95} & \textbf{\small{}75.83} & {\small{}94.23} & {\small{}94.31} & \textbf{\small{}93.27} & \textbf{\small{}91.33}\tabularnewline
\bottomrule 
\end{tabular}}
\end{table}

\begin{table}[t]
\centering{}\caption{\label{tab:CIFAR100eval}Performance in CIFAR-100 with
symmetric and asymmetric noise. ({*}) Denotes that we have run the algorithm.}
\medskip{}
\setlength{\tabcolsep}{3pt}\resizebox{1\columnwidth}{!}{{\small{}}%
\begin{tabular}{llllllll|l}
\toprule
 &  & \multicolumn{3}{l}{{\small{}Symmetric}} & \multicolumn{3}{l|}{{\small{}Asymmetric}} & {\small{}Avg.}\tabularnewline
 & {\small{}0\%} & {\small{}20\%} & {\small{}40\%} & {\small{}80\%} & {\small{}10\%} & {\small{}30\%} & {\small{}40\%} & \tabularnewline
\midrule 
{\small{}CE} & {\small{}74.34} & {\small{}58.75} & {\small{}42.92} & {\small{}8.29} & {\small{}68.10} & {\small{}53.28} & {\small{}44.46} & {\small{}50.02}\tabularnewline

{\small{}Mix \cite{2018_ICLR_mixup}} & {\small{}77.90} & {\small{}66.40} & {\small{}52.20} & {\small{}13.21} & {\small{}72.40} & {\small{}57.63} & {\small{}48.07} & {\small{}55.40}\tabularnewline
 
{\small{}DB \cite{2019_ICML_DynamicBootstrapping}} & {\small{}64.79} & {\small{}69.11} & {\small{}62.78} & {\small{}45.67} & {\small{}67.09} & {\small{}58.59} & {\small{}47.44} & {\small{}59.35}\tabularnewline
 
{\small{}DMI \cite{2019_NeurIPS_LDMI}} & {\small{}74.44} & {\small{}58.82} & {\small{}53.22} & {\small{}20.30} & {\small{}68.15} & {\small{}54.15} & {\small{}46.20} & {\small{}53.61}\tabularnewline

{\small{}PCIL \cite{2019_CVPR_JointOptimizImproved}} & {\small{}77.75} & {\small{}74.93} & {\small{}68.49} & {\small{}25.41} & {\small{}76.05} & {\small{}59.29} & {\small{}48.26} & {\small{}61.45}\tabularnewline
 
{\small{}DRPL \cite{2020_ICPR_SSLnoise}} & {\small{}71.84} & {\small{}71.16} & {\small{}72.37} & \textbf{\small{}52.95} & {\small{}72.03} & {\small{}69.30} & {\small{}65.69} & {\small{}67.91}\tabularnewline

{\small{}DMix{*} \cite{2020_ICLR_DivideMix}} & {\small{}67.41}  & {\small{}71.39}  & {\small{}70.83} & {\small{}49.52}  & {\small{}69.53}  & {\small{}68.28}  & {\small{}50.99} & {\small{}63.99} \tabularnewline

{\small{}ELR{*} \cite{2020_NeurIPS_EarlyReg}} & \textbf{\small{}78.01}  & \textbf{\small{}75.90}  & \textbf{\small{}72.89}  & {\small{}36.83}  & {\small{}77.08}  & {\small{}74.61}  & {\small{}71.25}  & {\small{}69.51} \tabularnewline
\midrule 
{\small{}MOIT} & {\small{}75.83} & {\small{}72.78} & {\small{}67.36} & {\small{}45.63} & {\small{}75.49} & {\small{}73.34} & {\small{}71.55} & {\small{}68.85}\tabularnewline

{\small{}MOIT+} & {\small{}77.07} & {\small{}75.89} & {\small{}70.88} & {\small{}51.36} & \textbf{\small{}77.43} & \textbf{\small{}75.13} & \textbf{\small{}74.04} & \textbf{\small{}71.69}\tabularnewline
\bottomrule
\end{tabular}}{\small{}}
\end{table}

\subsection{Web label noise evaluation}

\begin{table}[t]
\begin{centering}
\caption{\label{tab:miniImageNetEval}Performance evaluation on controlled
web noise in mini-ImageNet. We run all methods.}
\medskip{}
{\small{}}{\small\par}
\par\end{centering}
\centering{}{\small{}}%
\begin{tabularx}{\columnwidth}{Xlllll}
\toprule 
 & & {\small{}0\%} & {\small{}20\%} & {\small{}40\%} & {\small{}80\%}\tabularnewline
\midrule 
{\small{}Mix \cite{2018_ICLR_mixup}} & {\small{}Best} & {\small{}61.18} & {\small{}57.76} & {\small{}52.88} & {\small{}38.32}\tabularnewline
 & {\small{}Last} & {\small{}58.96} & {\small{}54.60} & {\small{}50.40} & {\small{}37.32} \tabularnewline

{\small{}DMix \cite{2020_ICLR_DivideMix}} & {\small{}Best} & {\small{}57.80} & {\small{}55.86} & {\small{}55.44} & {\small{}41.12}\tabularnewline
 & {\small{}Last} & {\small{}55.84} & {\small{}50.30} & {\small{}50.94} & {\small{}35.42} \tabularnewline

{\small{}ELR \cite{2020_NeurIPS_EarlyReg}} & {\small{}Best} & {\small{}63.12} & {\small{}61.48} & {\small{}57.32} & {\small{}41.68}\tabularnewline
 & {\small{}Last} & {\small{}57.38} & {\small{}58.10} & {\small{}50.62} & {\small{}41.68} \tabularnewline

\midrule
{\small{}MOIT} & {\small{}Best} & {\small{}67.18} & {\small{}64.82} & {\small{}61.76} & {\small{}46.40}\tabularnewline
 & {\small{}Last} & {\small{}64.72} & {\small{}63.14} & {\small{}60.78} & {\small{}45.88} \tabularnewline

{\small{}MOIT+} & {\small{}Best} & \textbf{\small{}68.28} & \textbf{\small{}64.98} & \textbf{\small{}62.36} & \textbf{\small{}47.80}\tabularnewline
 & {\small{}Last} & {\small{}67.82} & {\small{}63.10} & {\small{}61.16} & {\small{}46.78} \tabularnewline

\bottomrule
\end{tabularx}{\small\par}
\end{table}

\begin{table}[t]
\begin{centering}
\caption{\label{tab:WebVisionEval}Performance evaluation in mini-WebVision. We run all methods.}
\medskip{}
{\small{}}{\small\par}
\par\end{centering}
\centering{}{\small{}}%
\begin{tabularx}{\columnwidth}{llllll}
\toprule 

 & {\small{}Mix \cite{2018_ICLR_mixup}} & {\small{}DMix \cite{2020_ICLR_DivideMix}} & {\small{}ELR \cite{2020_NeurIPS_EarlyReg}} & {\small{}MOIT} & {\small{}MOIT+}\tabularnewline
\midrule 
{\small{}Best} & {\small{}74.96} & {\small{}76.08} & {\small{}73.00} & {\small{}78.36} & \textbf{\small{}78.76}\tabularnewline

{\small{}Last} & {\small{}73.76} & {\small{}74.64} & {\small{}71.88} & {\small{}77.76} & \textbf{\small{}78.72}\tabularnewline

\bottomrule 
\end{tabularx}{\small\par}
\end{table}
Tables \ref{tab:miniImageNetEval} and \ref{tab:WebVisionEval} illustrate the superior performance of MOIT/MOIT+ when training in the presence of web label noise in mini-ImageNet \cite{2020_ICML_DatasetOOD} and mini-WebVision \cite{2020_ICLR_DivideMix}. The results demonstrate that MOIT/MOIT+ are robust to web noise and that they do not need careful re-parametrization depending on the noise level or distribution to achieve state-of-the-art performance. The results in Tab. \ref{tab:miniImageNetEval} further confirm that the improvements are consistent across noise levels. It is interesting to observe that, although MOIT+ consistently outperforms MOIT, the improvements compared to CIFAR experiments tend to be smaller. We think that a plausible explanation is the dominance of out-of-distribution samples in web-noise, which makes label correction via semi-supervised learning less beneficial.
Note that we run M, DMix, EReg, and MOIT for the same number of epochs (130) in both mini-ImageNet and mini-WebVision.

\section{Conclusion}

This paper proposes Multi-Objective Interpolation Training (MOIT), an approach for image classification with deep neural networks that robustly learns in the presence of both synthetic and web label noise. The key idea of MOIT is to combine supervised contrastive learning and classification in such a way that they are both robust to label noise. Interpolated Contrastive Learning regularization enables learning label noise robust representations that are used to estimate a soft-label distribution whose agreement with
the original label allows identification of correctly labeled samples. MOIT then treats the remaining samples as unlabeled and trains a label noise robust image classifier in a semi-supervised manner. We further propose MOIT+, a refinement of our model by fine-tuning the model while re-training the image classifier. We conduct experiments in CIFAR-10/100 with synthetic label noise and in mini-ImageNet and mini-WebVision with web noise to demonstrate that MOIT and MOIT+ achieve state-of-the-art results when training deep neural networks with different noise distributions and levels. Future work will explore instance-dependent label noise as well as how to simplify the contrastive learning framework by using class prototypes.

\subsection*{Acknowledgements}
This publication has emanated from research conducted with the financial support of Science Foundation Ireland (SFI) under grant number SFI/15/SIRG/3283 and SFI/12/RC/2289 P2.

{\small{}{}{} \bibliographystyle{ieee_fullname}
\bibliography{refs}
 } 
\end{document}